\def\year{2018}\relax
\documentclass[letterpaper]{article} 
\usepackage{aaai18}  
\usepackage{times}  
\usepackage{helvet}  
\usepackage{courier}  
\usepackage{url}  
\usepackage{graphicx}  
\usepackage{algorithm,algorithmic}

\usepackage{multirow}
\usepackage{multicol}
\usepackage{kotex}
\usepackage{balance}
\usepackage{color}

\usepackage[cmex10]{amsmath}
\usepackage{amssymb}

\begin{document}
%
\title{Less-forgetful Learning for Domain Expansion in Deep Neural Networks}

\author{Heechul Jung\\
School of EE\\
KAIST\\
heechul@kaist.ac.kr\\
\And
Jeongwoo Ju\\
Division of Future Vehicle\\
KAIST\\
veryju@kaist.ac.kr\\
\And
Minju Jung\\
School of EE\\
KAIST\\
alswn0925@kaist.ac.kr\\
\And
Junmo Kim\\
School of EE\\
KAIST\\
junmo.kim@kaist.ac.kr\\
}
\maketitle
\begin{abstract}
Expanding the domain that deep neural network has already learned without accessing old domain data is a challenging task because deep neural networks forget previously learned information when learning new data from a new domain. In this paper, we propose a less-forgetful learning method for the domain expansion scenario. While existing domain adaptation techniques solely focused on adapting to new domains, the proposed technique focuses on working well with both old and new domains without needing to know whether the input is from the old or new domain. First, we present two naive approaches
which will be problematic, then we provide a new method using two proposed properties for less-forgetful learning. Finally, we prove the effectiveness of our method through experiments on image classification tasks. All datasets used in the paper, will be released on our website for someone’s follow-up study.
\end{abstract}

\section{Introduction}
\label{sec:int}

\begin{figure}[t!]
\begin{center}
\includegraphics[width=\linewidth]{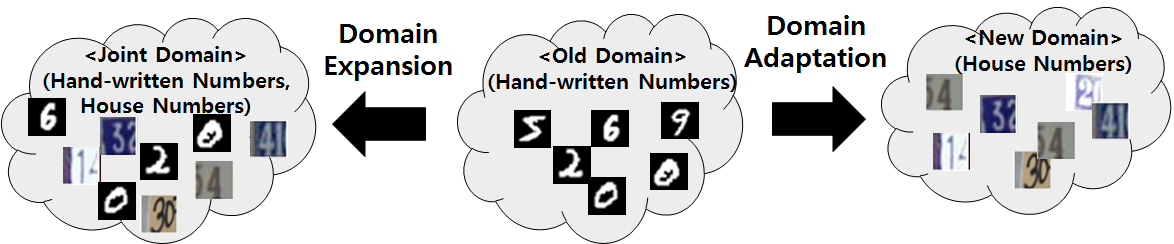}\\
\small
\caption{\textbf{Domain expansion and domain adaptation.} Domain expansion enables DNNs to adapt to new environments while preserving their performance within old environments. Joint domain involves both old and new domains.}
\label{fig:concept}
\end{center}
\vspace{-3mm}
\end{figure}

Deep neural networks (DNNs) have advanced to nearly human levels of object, face, and speech recognition \cite{taigman2014deepface} \cite{graves2013speech} \cite{szegedy2014going} \cite{simonyan2014very} \cite{zhang2016joint} \cite{richardson2015deep}. Despite these advances, issues still remain. Domain adaptation (the same tasks but in different domains) is one of these remaining issues \cite{ganin2014unsupervised} \cite{ganin2016domain} \cite{long2015learning}. The domain adaptation problem concerns how well a DNN works in a new domain that has not been learned. In other words, these domain adaptation techniques focus on adapting only to new domains, but in an actual situation, applications often need to remember old domains as well without seeing the old domain data again. We call this the DNN domain expansion problem. Its concept is illustrated in Figure \ref{fig:concept}.

For example, suppose you have an object recognition system mounted on a robot or a smartphone that has been trained with ImageNet dataset \cite{ILSVRC15}. The real-world environment is so diverse (e.g., with various lighting changes) that the system will sometimes fail. Learning the failed data collected from the real-world environment might prevent the repetition of the failure when the DNN encounters the same situation. Unfortunately, the DNN forgets the information previously learned from ImageNet dataset while learning the failed data collected from the real-world environment. In other words, the object recognition system gradually loses its original ability, and hence, requires the domain expansion functionality to preserve its ability for the ImageNet domain and adapt to the new domain that was not covered by the ImageNet dataset.


The DNN domain expansion problem is specifically important for the following three main reasons:
\begin{itemize}
\item It enables the DNNs to continually learn from sequentially incoming data.
\item In practice, users can fine-tune their DNNs using only new data collected from new environments without access to data from the old domain.
\item Making a single unified network that performs in several domains is possible.
\end{itemize}

In this paper, we propose a method to enable DNNs to achieve domain expansion functionality by alleviating the forgetting problem.

\section{Domain Expansion Problem}
\label{sec:de}
\begin{figure}[t!]
\begin{center}
\includegraphics[width=2.3cm]{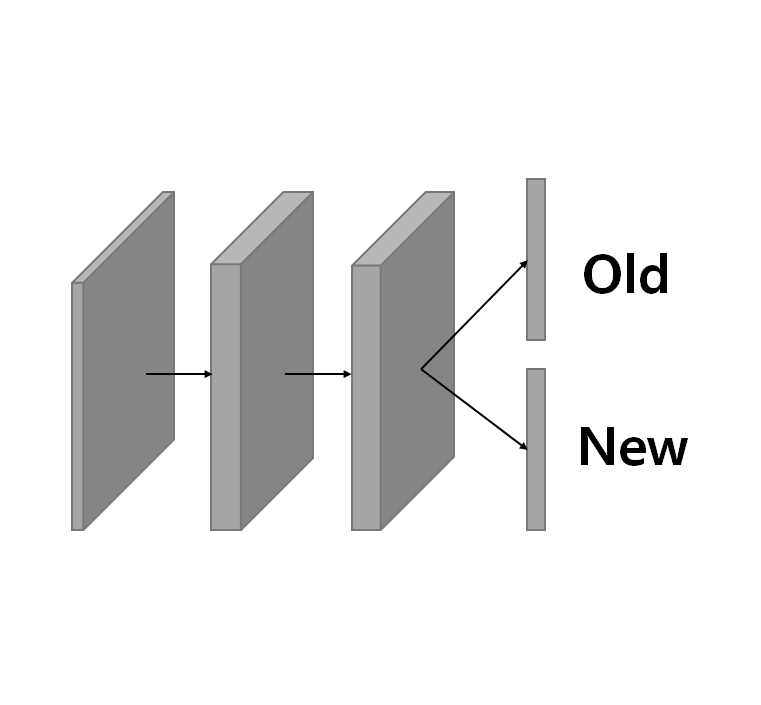}\hspace{0.7cm}\includegraphics[width=2.3cm]{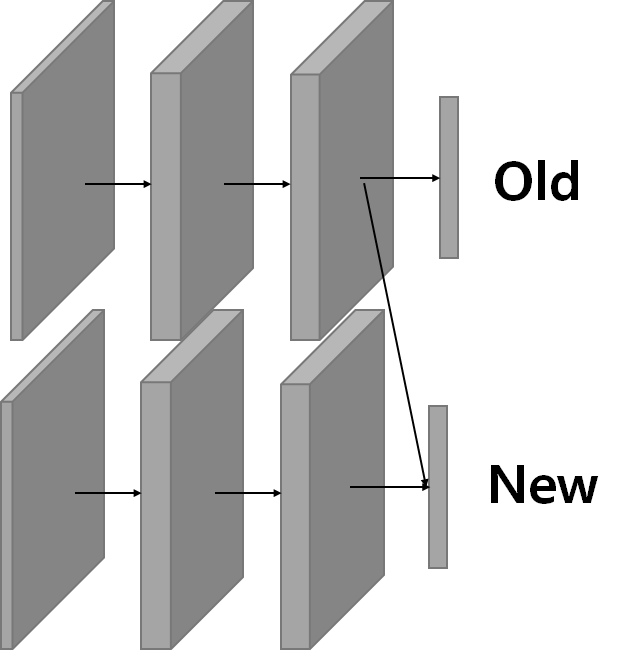}\hspace{0.7cm}\includegraphics[width=2.3cm]{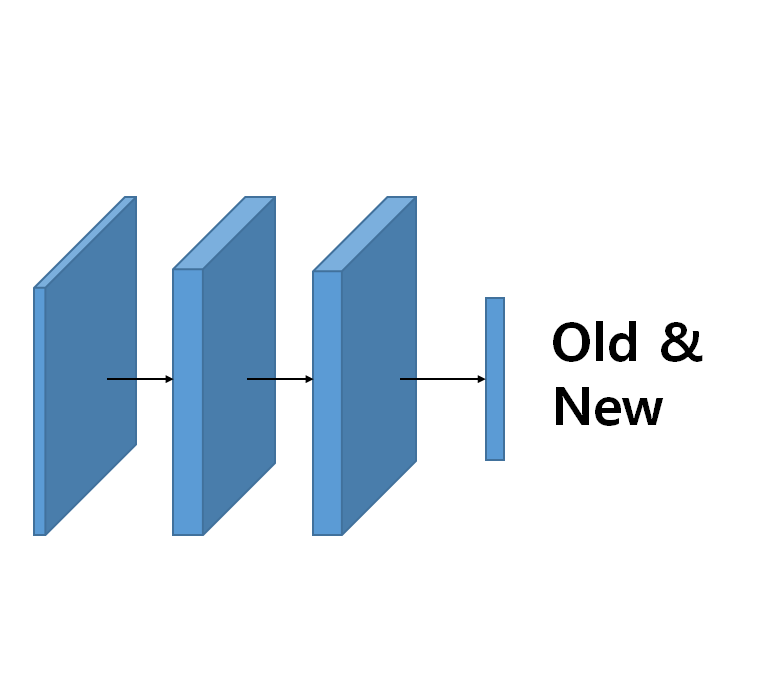}\\
\footnotesize
(a) Type A\hspace{15mm}(b) Type B\hspace{15mm}(c) Type C
\vspace{-3mm}
\end{center}
\small
\caption{\textbf{Three types of various learning techniques that use information from both old and new tasks or domains together.} (a) LwF \cite{li2016learning} (b) Progressive learning \cite{rusu2016progressive}. (c) Less-forgetful. The existing methods (a) and (b) need to know in advance whether the input data come from the old domain or the new domain. In contrast, our method (c) does not need this prior knowledge.}
\label{fig:methods}
\vspace{-3mm}
\end{figure}

We define the \textbf{domain expansion problem} as the problem of creating a network that works well both on an old domain and a new domain even after it is trained in a supervised way using only the data from the new domain without accessing the data from the old domain.
Two challenging issues need to be faced in solving the domain expansion problem.
First, the performance of the network on the old domain should not be degraded even if the new domain data are learned without seeing those of the old domain (A general term is the \textit{catastrophic forgetting problem}).
Second, a DNN should work well without any prior knowledge of which domain the input data had come from. Figures \ref{fig:methods} (a) and (b) show the existing techniques that preserve the ability for old domain, but require prior knowledge about the data domain. Figure \ref{fig:methods} (c) shows our proposed method that preserves the old domain and does not require prior knowledge about the input data. Therefore, we focus on developing a new method to alleviate the catastrophic forgetting problem without any prior knowledge (e.g. old or new domain) about the input data.

Actually, the domain expansion problem is a part of the continual learning problem. The continual learning generally considers multiple task learning or sequence learning (more than two domains), whereas the domain expansion problem only considers two domains such as old and new. 

\section{Related Work}
In this section we will list the state-of-the-art techniques for solving the catastrophic forgetting problem.
Srivastava et al. proposed a local winner-take-all (LWTA) activation function that helps to prevent the forgetting problem \cite{srivastava2013compete}. This activation function is effective because it implements implicit long-term memory.
Subsequently, several experiments on the forgetting problem in the DNNs were empirically performed in \cite{goodfellow2013empirical}. The results showed that a dropout method \cite{Hinton2012} \cite{srivastava2014dropout} with a maxout \cite{goodfellow2013maxout} activation function was helpful in forgetting less of the learned information.
In addition, \cite{goodfellow2013empirical} stated that a large DNN with a dropout method can address the catastrophic forgetting problem.

An unsupervised approach was also proposed in \cite{goodrich2014unsupervised}. Goodrich et al. extended this method to a recurrent neural network \cite{goodrichmitigating2015}. These methods used an online clustering method that can help mitigate forgetting in a data-driven manner.
These methods computed cluster centroids while learning the training data in the old domain and using the computed centroids for the new domain. 


The learning without forgetting (LwF) method
 \cite{li2016learning} was also proposed to improve the DNN performance in a new task (Figure \ref{fig:methods} (a)). This method utilizes the knowledge distillation loss method to maintain the performance on the old data.
Google DeepMind \cite{rusu2016progressive} proposed a unified DNN based on progressive learning (PL) (Figure \ref{fig:methods} (b)). The PL method enables one network to operate several tasks. (The applications in \cite{rusu2016progressive} were Atari and three-dimensional maze games.) The idea is to use previously learned features when performing a new task via lateral connections. As mentioned in Section \ref{sec:de}, these methods are difficult to directly apply to the domain expansion problem without any modification because they need to know information about the input data domain.

Elastic weight consolidation (EWC) is one of the methods used to solve the catastrophic forgetting problem \cite{kirkpatrick2017overcoming}. This technique uses a Fisher information matrix computed from the old domain training data, and uses its diagonal elements as coefficients of $l_2$ regularization to obtain similar weight parameters between the old and new networks when learning the new domain data. 
Furthermore, generative adversarial networks are also used for generating old domain data while learning new domain data \cite{shin2017continual}.
\begin{table}[h!]
\renewcommand{\arraystretch}{1.2}
\centering
\scriptsize
\caption{Different types of state-of-the-art methods.}
\label{tb:types}
\begin{center}
\begin{tabular}{c|ccc}
\hline\hline
Type & Type A & Type B & Type C\\\hline
Type A$^\prime$ & - & - & EwC, ReplayGAN\\
Type B$^\prime$ & LwF & PL & \textbf{Proposed Method} \\
\hline\hline
\end{tabular}
\end{center}
\vspace{-3mm}
\end{table}

\begin{figure}[t!]
\begin{center}
\includegraphics[width=4.3cm]{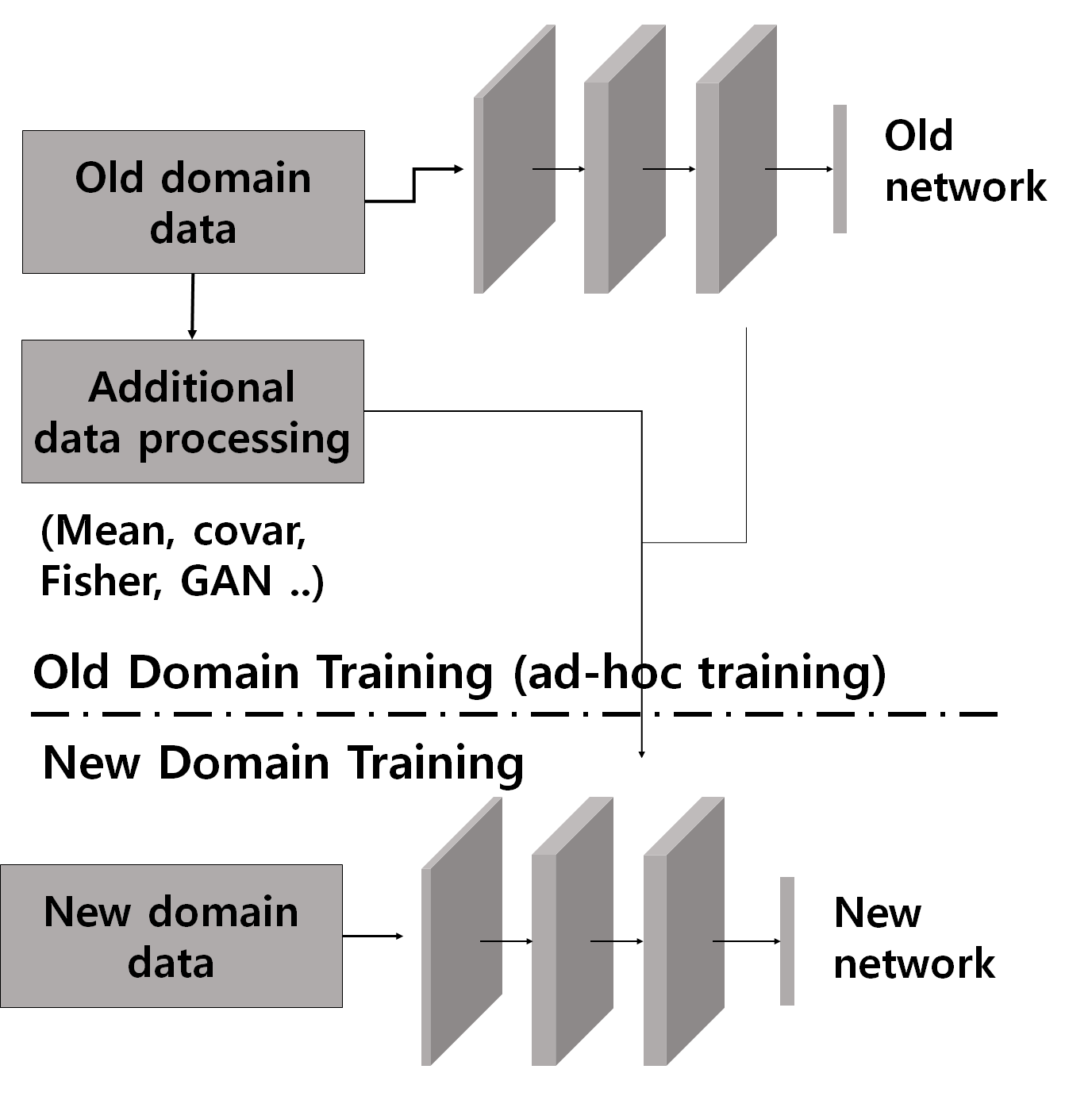}\includegraphics[width=4.3cm]{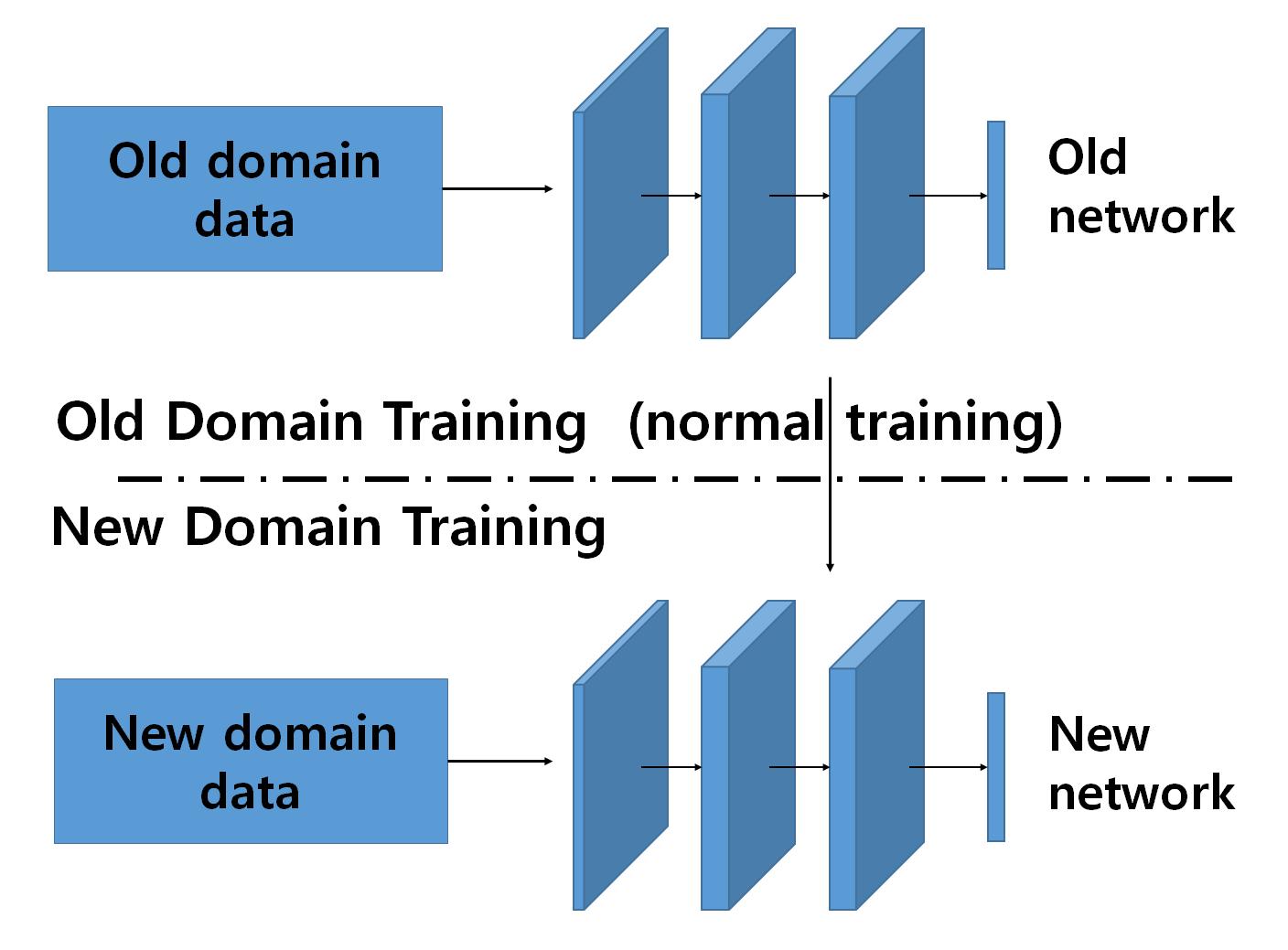}\\
\footnotesize
(a) Type A$^\prime$\hspace{28mm}(b) Type B$^\prime$
\vspace{-3mm}
\end{center}
\small
\caption{\textbf{Two types of training processes on old domain to alleviate catastrophic forgetting problem.} (a) Ad-hoc training for old-domain. (b) Normal training for old-domain.}
\label{fig:types}
\vspace{-3mm}
\end{figure}

State-of-the-art algorithms can be classified into two types, as shown in Figure \ref{fig:types}. The algorithms shown in Figure \ref{fig:types} (a) go through an ad-hoc training process to extract useful information from the old domain data. The information extracted from the old domain data will be used to alleviate catastrophic forgetting problem when the network learns new domain data. Figure \ref{fig:types} (b) shows the proposed method; our method uses the usual way to train the network using old domain data. This gives a benefit that our method can be directly applied to any pre-trained models that can be downloaded from the Internet, without access to the old domain training data. Table \ref{tb:types} summarizes state-of-the-art algorithms for each type shown in Figures \ref{fig:methods} and \ref{fig:types}.

\section{Reformulation of Forgetting Problem}
\label{sec:lfp}
We denote the dataset for the old domain as $\mathrm{D}^{(o)} = \{(x_i^{(o)},y_i^{(o)})\}_{i=1}^{N_o}$ and the dataset for the new domain as $\mathrm{D}^{(n)} = \{(x_i^{(n)},y_i^{(n)})\}_{i=1}^{N_n}$, where $N_o$ and $N_n$ are the number of data points of the old and new domains, respectively. Furthermore, $x_i^{(\cdot)}$ is the training data, and $y_i^{(\cdot)}$ is the corresponding label. These two datasets are mutually exclusive. Each dataset has both the following training and validation datasets:
$\mathrm{D}^{(o)} = \mathrm{D}_t^{(o)} \cup \mathrm{D}_v^{(o)},
\mathrm{D}_t^{(o)} \cap \mathrm{D}_v^{(o)}=  \varnothing, \mathrm{D}^{(n)} = \mathrm{D}_t^{(n)} \cup \mathrm{D}_v^{(n)}$, and $\mathrm{D}_t^{(n)} \cap \mathrm{D}_v^{(n)}=  \varnothing, $
where $\mathrm{D}_t^{(\cdot)}$ and $\mathrm{D}_v^{(\cdot)}$ are the training and validation datasets, respectively.

The old network $\mathrm{F}(x; \theta^{(o)})$ for the old domain is trained using $\mathrm{D}_t^{(o)}$, where $\theta^{(o)}$ is a weight parameter set for the old domain. The initial values of the weights are randomly initialized using normal distribution $\mathcal{N}(0,\sigma^2)$.
The trained weight parameters $\theta^{(o)}$ for the old domain are obtained using dataset $\mathrm{D}_t^{(o)}$.
The new network $\mathrm{F}(x; \theta^{(n)})$ for the expanded domain, which is union of the old domain and the new domain, is trained using dataset $\mathrm{D}_t^{(n)}$ without access to the old domain training data $\mathrm{D}_t^{(o)}$. 
Finally, we obtain the updated weight parameters $\theta^{(n)}$ to satisfy the \textit{less-forgetful condition}, $\mathrm{F}(x;\theta^{(n)}) \approx \mathrm{F}(x;\theta^{(o)})$ for $x$ from $\mathrm{D}^{(o)}$. Our goal is to develop a method to satisfy the condition.

\section{Naive Approach}
\label{sec:naive}
\subsection{Fine-tuning only the softmax classifier layer}
\label{naive_fine}
The most common method to use, such that the DNN does not forget what it has learned, is to freeze lower layers and fine-tune the final softmax classifier layer. This method regards the lower layer as a feature extractor and updates the linear classifier to adapt to new domain data. In other words, the feature extractor is shared between the old and new domains, and the method seems to preserve the old domain information.

\subsection{Weight constraint approach}
\label{sec:weight}
The weight constraint method is a method that uses $l_2$ regularization to obtain similar weight parameters between the old and new networks when learning the new data as follows:
\begin{equation}
\mathcal{L}_{w}(x;\theta^{(o)}, \theta^{(n)}) = \lambda_c\mathcal{L}_c(x;\theta^{(n)}) + \lambda_w||\theta^{(o)}-\theta^{(n)}||_2,
\label{eq:weight_constraint}
\end{equation}
where $\lambda_c$ and $\lambda_w$ control the weight of each term, and $x$ comes from $\mathrm{D}^{(n)}$. The cross-entropy loss $\mathcal{L}_c$ is defined as follows:
\begin{equation}
\mathcal{L}_{c}(x;\theta^{(n)}) = -\sum_{i=1}^{C}t_i \log (o_i(x;\theta^{(n)})),
\end{equation}
where $t_i$ is the $i$-th value of the ground truth label; $o_i$ is the $i$-th output value of the softmax of the network; and $C$ is the total number of classes. The parameter $\theta^{(n)}$ is initialized to $\theta^{(o)}$. We then compute the new weight parameter $\theta^{(n)}$ by minimizing the loss function $\mathcal{L}_{w}$. This method was designed with the expectation that the learned information will be preserved if the weight parameter does not change much.

\section{Less-forgetful learning}
In general, the lower layer in DNNs is considered as a feature extractor, while the top layer is regarded as a linear classifier, which means that the weights of the softmax classifier represent a decision boundary for classifying the features.
The features extracted from the top hidden layer are usually linearly separable because of the linear nature of the top layer classifier.
Using this knowledge, we propose a new learning scheme that satisfies the following two properties to reduce the tendency of the DNN to forget information learned from the old domain: \\

\noindent\textbf{Property 1.} \textit{The decision boundaries should be unchanged.}\\\\
\noindent\textbf{Property 2.} \textit{The features extracted by the new network from the data of the old domain should be present in a position close to the features extracted by the old network from the data of the old domain.}\\

 
\begin{figure}[b]
\begin{center}
\includegraphics[width=8cm]{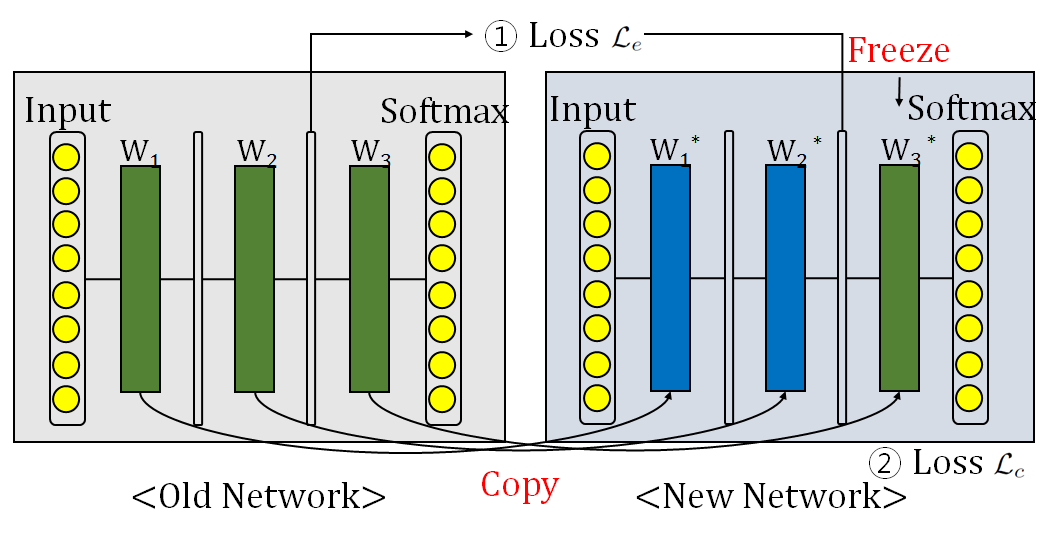}
\vspace{-5mm}
\caption{\textbf{Less-forgetful learning method.} Our learning method uses the trained weights of the old network as the initial weights of the new network and simultaneously minimizes two loss functions.}
\label{fig:overview}
\end{center}
\end{figure}
We build the less-forgetful learning algorithm based on these two properties. The first property is easily implemented by setting the learning rates of the boundary to zero. However, satisfying the second property is not trivial because we cannot access the old domain data. Therefore, instead of using the old domain data, we use the training data of the new domain and show that it is also helpful in satisfying Property 2.  

Figure \ref{fig:overview} briefly shows our algorithm. The details of which are as follows: as in the traditional fine-tuning method, we initially reuse the weights of the old network, which was trained using the training data of the old domain, as the initial weights of the new network. Next, we freeze the weights of the softmax classifier layer to preserve the boundaries of the classifier, then we train the network to minimize the total loss function as follows:
\begin{equation}
\mathcal{L}_{t}(x;\theta^{(o)}, \theta^{(n)}) = \lambda_c\mathcal{L}_c(x;\theta^{(n)}) + \lambda_e\mathcal{L}_e(x;\theta^{(o)}, \theta^{(n)}),
\label{eq:total_loss}
\end{equation}
where $\mathcal{L}_t$, $\mathcal{L}_c$, and $\mathcal{L}_e$ are the total, cross-entropy, and Euclidean loss functions, respectively; $\lambda_c$ and $\lambda_e$ are the tuning parameters for adjusting the scale between the two loss values; and $x$ comes from $\mathrm{D}^{(n)}$. Parameter $\lambda_e$ usually has a smaller value than $\lambda_c$. $\lambda_c$ is set to one for all the experiments in this paper.

The cross-entropy loss function $\mathcal{L}_c$ defined in Eq. (2) helps the network to correctly classify input data $x$.
$\mathcal{L}_e$ is defined as follows to satisfy the proposed second property:
\begin{equation}
\mathcal{L}_{e}(x;\theta^{(o)}, \theta^{(n)}) = \frac{1}{2} ||\mathbf{f}_{L-1}(x;\theta^{(o)})-\mathbf{f}_{L-1}(x;\theta^{(n)})||^2_2,
\end{equation}
where $L$ is the total number of layers, and $\mathbf{f}_{L-1}$ is a feature vector of layer $L-1$, which is just before the softmax classifier layer. The new network learns to extract features similar to the features extracted by the old network using the loss function. 
We obtain the following equation:
\begin{equation}
\hat{\theta}^{(n)} = \arg\min_{\theta^{(n)}}\mathcal{L}_{t}(x;\theta^{(o)}, \theta^{(n)}) + \mathcal{R}(\theta^{(n)}),
\label{eq:final_loss}
\end{equation}
where $\mathcal{R}(\cdot)$ denotes a general regularization term, such as weight decay.
\begin{algorithm}[t]
\footnotesize
\caption{Less-forgetful (LF) learning}
\textbf{Input:} $\theta^{(o)}, \mathrm{D}_t^{(n)}, N, N_b$\\
\textbf{Output:} $\hat{\theta}^{(n)}$
\begin{algorithmic}[1]
  \STATE $\theta^{(n)}\leftarrow\theta^{(o)}$ // initial weights
  \STATE Freeze the weights of the softmax classifier layer.
  \STATE for i=1,$\ldots$,$N$ // training iteration
  \STATE ~~~Select mini-batch set $\mathrm{B}$ from $\mathrm{D}_t^{(n)}$, where $|\mathrm{B}| = N_b$.
  \STATE ~~Update $\theta^{(n)}$ using backpropagation with $\mathrm{B}$ to minimize total loss $\mathcal{L}_t(x;\theta^{(o)}, \theta^{(n)})  + \mathcal{R}(\theta^{(n)})$.
  \STATE end for
  \STATE $\hat{\theta}^{(n)} \leftarrow \theta^{(n)}$
  \STATE Return $\hat{\theta}^{(n)}$
\end{algorithmic}
\end{algorithm}
Finally, we build the less-forgetful learning algorithm, as shown in Algorithm 1. Parameters $N$ and $N_b$ in the algorithm denote the number of iterations and the size of mini-batches, respectively.
\section{Experimental results}
\label{sec:exp}
\subsection{Details of Datasets}
\begin{figure}[t!]
\begin{center}
\includegraphics[width=1.3cm]{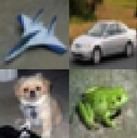}
\includegraphics[width=1.3cm]{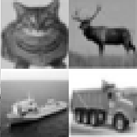}
\includegraphics[width=1.3cm]{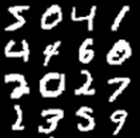}
\includegraphics[width=1.3cm]{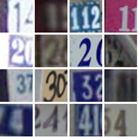}
\includegraphics[width=1.3cm]{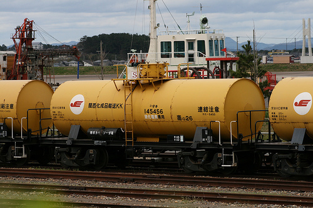}
\includegraphics[width=1.3cm]{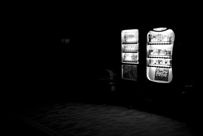}
\end{center}
\vspace{-3mm}
\caption{\textbf{Example images of the datasets used in the experiments.} From left to right: CIFAR Color $\cup$ CIFAR Gray, MNIST $\cup$ SVHN, and ImageNet Normal $\cup$ ImageNet Dark \& Bright.}
\label{fig:dataset}
\vspace{-3mm}
\end{figure}

We conducted two different experiments for image classification: one using datasets consisting of tiny images (CIFAR-10 \cite{Krizhevsky2009Learning}, MNIST \cite{LeCun1998Gradient}, SVHN \cite{Netzer2011Reading}) and one using a dataset made up of large images (ImageNet \cite{ILSVRC15}). Figure \ref{fig:dataset} shows example images from the datasets that we used in the experiments. Table \ref{tb:db_num} presents the number of images for each dataset. The original training and test data for the SVHN dataset were 73,257 and 26,032, respectively. However, we randomly selected some images in the dataset to match the number of images with those of the MNIST dataset. 
\begin{table}[t!]
\renewcommand{\arraystretch}{1.2}
\centering
\scriptsize
\caption{Number of images for each dataset used in the experiments.}
\label{tb:db_num}
\begin{center}
\begin{tabular}{c|ccc}
\hline\hline
Old domain & MNIST & CIFAR-10 Color& ImageNet Normal\\ \hline
Train & 60,000 & 40,000 & 52,503\\
Test & 10,000 & 10,000 & 5,978 \\ \hline
New domain& SVHN & CIFAR-10 Gray& ImageNet Dark \& Bright\\ \hline
Train & 60,000 & 10,000 & 4,445\\
Test & 10,000 & 10,000 & 505 \\
\hline\hline
\end{tabular}
\end{center}
\vspace{-3mm}
\end{table}

\subsection{Details of Comparison Methods}
Next, we compare the classification performance of the proposed algorithm with that of the state-of-the-art methods.
First, we test two naive approaches, weight constraint and fine-tuning, on the softmax classifier layer (Fine-tuning (Linear)), and we use this as the baseline. Fine-tuning with various activation functions such as ReLU, Maxout, \cite{goodfellow2013empirical} and LWTA \cite{srivastava2013compete} are also used for performance comparison. Further, we show classification rates of recent works such as LwF \cite{li2016learning} and EWC \cite{kirkpatrick2017overcoming}. 
\subsection{Implementation Detail}
We used the Caffe framework for implementing our algorithm and baseline methods \cite{jia2014caffe}.
Architectures for the tiny image classification experiment are shown in Table \ref{tb:architec}. Three consecutive convolutional layers and a fully connected layer were used with ReLU or Maxout or LWTA, and the last softmax classifier layer comprised of $10$ nodes.
We used GoogleNet \cite{szegedy2014going} as the ImageNet dataset, and the number of nodes of the softmax classifier layer was set to $50$. 
Parameters for the solvers are listed as in Table \ref{tb:solver}. All the experiments such as fine-tuning, weight constraint, modified LwF, and LF were implemented using the same parameters and architectures.

\begin{table}[h!]
\centering
\scriptsize
\caption{Architecture details of DNNs for each dataset}
\vspace{2mm}
\label{tb:architec}
\begin{tabular}{c|c|c}
\hline\hline
Dataset                 & MNIST $\cup$ SVHN         & CIFAR-10 COLOR $\cup$ GRAY \\ \hline
\multirow{15}{*}{Layers} & INPUT (28$\times$28$\times$3) & INPUT (32$\times$32$\times$3) \\ \cline{2-3} 
                      & CONV (5$\times$5$\times$32)  & CONV(5$\times$5$\times$32) \\ 
                        & ReLU or Maxout or LWTA         & ReLU or Maxout or LWTA          \\ \cline{2-3} 
                        & MAXPOOL (3$\times$3,2)& MAXPOOL (3$\times$3,2)  \\ \cline{2-3} 
                        & CONV(5$\times$5$\times$32)  & CONV(5$\times$5$\times$32) \\ 
                        & ReLU or Maxout or LWTA         & ReLU or Maxout or LWTA          \\ \cline{2-3} 
                        & MAXPOOL (3$\times$3,2)& MAXPOOL (3$\times$3,2)        \\ \cline{2-3} 
                        & CONV(5$\times$5$\times$64)  & CONV (5$\times$5$\times$64)         \\ 
                        & ReLU or Maxout or LWTA          & ReLU or Maxout or LWTA        \\ \cline{2-3} 
                        & MAXPOOL (3$\times$3,2)& MAXPOOL (3$\times$3,2)          \\ \cline{2-3} 
                        & FC (200)       & FC (200)       \\ 
                        & ReLU or Maxout or LWTA         & ReLU or Maxout or LWTA          \\ \cline{2-3} 
                        & FC (10)        & FC (10)        \\ \cline{2-3} 
                        & SOFTMAX   & SOFTMAX           \\ \hline\hline
\end{tabular}
\end{table}

\begin{table}[h]
\centering
\scriptsize
\caption{Parameters used in experiments.}
\label{tb:solver}
\begin{tabular}{c|c|c|c|c}
\hline\hline
Exp. type & Tiny & Tiny & Realistic & Realistic\\\hline
Domain type & Old & New & Old & New \\
\hline
mini-batch size & 100 & 100 & 128 & 64\\
learning rate (lr) & 0.01 & 0.0001 & 0.01 & 0.001\\ 
lr policy& step & fix & step & fix\\ 
decay & 0.1 & -  & 0.1 & - \\ 
step size& 20000 & - & 20000 & -\\ 
max\_iter& 40000 & 10000  & 100000 & 1000\\ 
momentum& 0.9 & 0.9 & 0.9 & 0.9 \\ 
weight\_decay& 0.004 & 0.004 & 0.0005 & 0.0005 \\ \hline\hline
\end{tabular}
\vspace{-3mm}
\end{table}

\subsection{Tiny image classification (MNIST, SVHN, and CIFAR-10)}
 We built two experimental scenarios to evaluate our method using the tiny image datasets. The first scenario was the domain expansion from the MNIST to the SVHN (MNIST $\cup$ SVHN), while the second one was the domain expansion from the color to grayscale images using the CIFAR-10 dataset (CIFAR Color $\cup$ CIFAR Gray). We also compared the proposed method with various existing methods, such as traditional fine-tuning, fine-tuning only the softmax classifier layer (Linear), weight constraint method, and modified LwF, to demonstrate the superiority of our method. 
Please see the supplementary material for details on the modified LwF method.

\begin{table}[b!]
\centering
\scriptsize
\caption{Experimental results for the tiny dataset experiments.}
\label{tb:Result}
\begin{center}
\begin{tabular}{c|cccc}
\hline\hline
&Methods&Old (\%)&New (\%)&Avg. (\%)\\ \hline
& Old network (ReLU) & 99.32 & 31.04 & 65.14 \\
 & Old network (Maxout) & 99.50 & 29.07 & 64.29 \\
&  Old network (LWTA) & 99.50 & 27.50 & 63.50 \\ \cline{2-5} 
& Fine-tuning (ReLU) & 59.93 & 87.83 & 73.88 \\ 
& Fine-tuning (Linear) & 67.43 & 52.01 & 59.72\\ 
MNIST & Fine-tuning (Maxout) & 64.82 & 86.44 & {75.63} \\
$\cup$ & Fine-tuning (LWTA) & 58.38 & 82.80 & 70.59 \\
SVHN & Weight constraint & 80.29 & 86.60 & 83.45\\
 & Modified LwF ($\beta=0.5)$ & 94.78 & 83.77 & 89.28\\ 
 & EWC ($\gamma=2.32\times10^{4}$) & 94.15 & 79.31 & 86.73\\\cline{2-5} 
 & \textbf{LF} ($\lambda_{e}=1.6\times10^{-3}$) & {97.37} & {83.79} & \textbf{90.58}\\
 & \textbf{LF} ($\lambda_{e}=7.8\times10^{-4}$) & {95.18} & {85.93} & {90.56} \\
 & \textbf{LF} ($\lambda_{e}=3.9\times10^{-4}$) & {90.89} & {87.57} & {89.23} \\
 & \textbf{LF} ($\lambda_{e}=2.0\times10^{-4}$) & {85.27} & {88.55} & {86.91}\\ \hline
& Old network (ReLU) & 77.84 & 64.09 & 70.96 \\
& Old network (Maxout) & 78.64 & 64.90 & 71.77 \\
 & Old network (LWTA) & 76.04 & 65.72 & 70.88 \\ \cline{2-5} 
& Fine-tuning (ReLU) & 69.40 & 70.84 & 70.12 \\ 
CIFAR & Fine-tuning (Linear) & {73.85} & 71.95 & 72.90 \\
Color & Fine-tuning (Maxout) & 71.06 & 73.07 & {72.07} \\
$\cup$ & Fine-tuning (LWTA) & 68.21 & 72.99 & 70.60 \\ 
CIFAR & Weight constraint & 72.44 & 74.40 & 73.42 \\
Gray & Modified LwF ($\beta=3$) & 75.87 & 72.79 & 74.33 \\
& EWC ($\gamma=10^{4}$) & 75.56 & 72.21 & 73.89 \\ \cline{2-5} 
& \textbf{LF} ($\lambda_{e}=1.6\times10^{-3}$) & {75.83} & {73.70} & \textbf{74.77} \\ 
& \textbf{LF} ($\lambda_{e}=7.8\times10^{-4}$) & {74.75} & {74.60} & 74.68 \\ 
 & \textbf{LF} ($\lambda_{e}=3.9\times10^{-4}$) & {73.77} & {74.43} & 74.1 \\
 & \textbf{LF} ($\lambda_{e}=2.0\times10^{-4}$) & {72.71} & {74.31} & 73.51\\ \hline\hline
\end{tabular}
\end{center}
\vspace{-3mm}
\end{table}

Table \ref{tb:Result} shows the classification rates obtained by the test sets of each data set. The ``old network'' method in Table \ref{tb:Result} indicates the training using only the training data of the old domain. The rest of the table shows the results of further training using each method with the training data of the new domain. In addition, the columns ``old'' and ``new'' in Table \ref{tb:Result} represent the classification rates for each domain, while ``avg.'' represents the average of the two classification rates. $\beta$ and $\gamma$ in Table \ref{tb:Result} are hyper parameters for the modified LwF and EWC. $\beta$ is explained in supplementary material, and $\gamma$ denotes $\lambda$ in the original EWC paper \cite{kirkpatrick2017overcoming}.

Our method outperformed state-of-the art methods, such as the modified LwF and EWC. The method that only fine-tuned the linear classifier failed to adapt to the new domain because of only a few learnable parameters available to learn the new domain. Meanwhile, the weight constraint method forgot the old domain information much more than our method.

\begin{figure}[t!]
\begin{center}
\includegraphics[width=4cm]{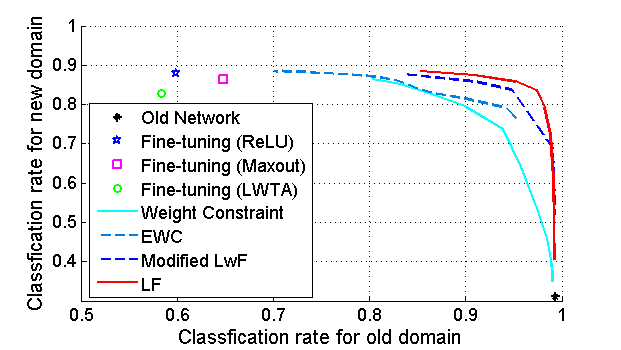}
\includegraphics[width=4cm]{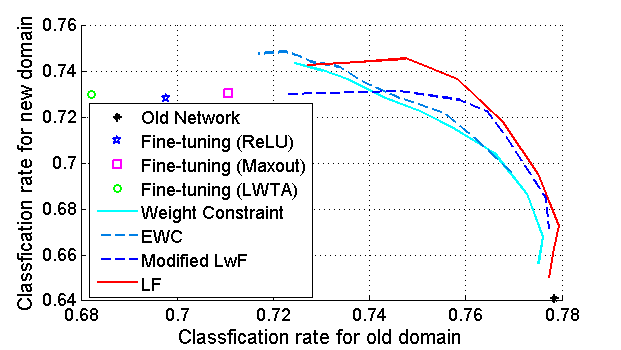}
\vspace{-3mm}
\caption{\textbf{Relationship between the classification rates for the old and new domains. \textit{[Best viewed in color]}} (Left) Results for the MNIST $\cup$ SVHN and (right) for the CIFAR Color $\cup$ CIFAR Gray. The curves were generated according to the different values of $\lambda_e$ used in Eq. \ref{eq:total_loss}.}
\label{fig:roc}
\end{center}
\vspace{-3mm}
\end{figure}

\begin{figure}[t!]
\begin{center}
\includegraphics[width=4cm]{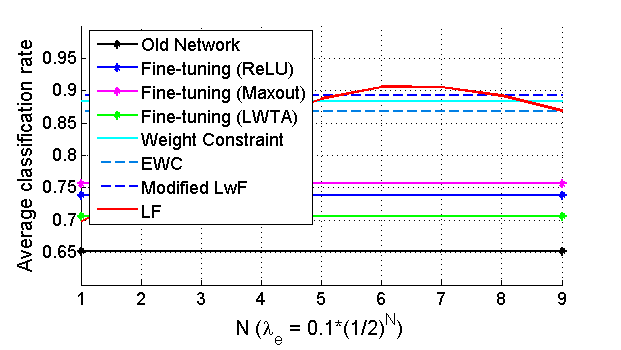}
\includegraphics[width=4cm]{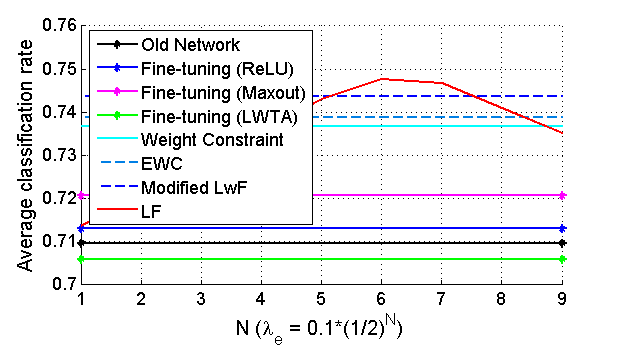}
\vspace{-3mm}
\caption{\textbf{Average classification rates with respect to $\lambda_e$. \textit{[Best viewed in color]}} (Left) MNIST $\cup$ SVHN (Right) CIFAR Color $\cup$ CIFAR Gray. For other algorithms, we got cherry picking to show their best performance.}
\label{fig:avg}
\end{center}
\vspace{-3mm}
\end{figure}

\begin{figure}[t!]
\begin{center}
\includegraphics[width=6cm]{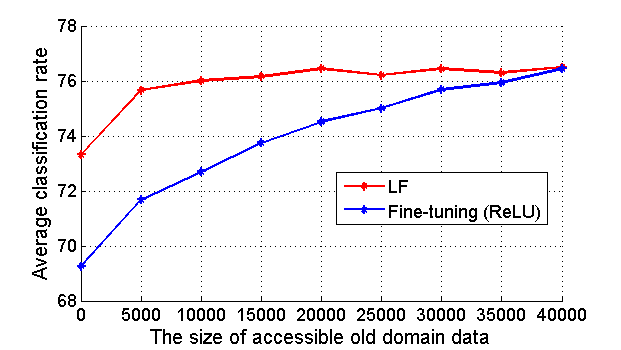}\\
\vspace{-3mm}
\caption{\textbf{Classification rates according to the size of the old-domain data using the CIFAR-10 dataset. \textit{[Best viewed in color]}} LF shows better classification rates than a traditional fine-tuning method when the number of accessible old-domain data is small.}
\label{fig:numofsource}
\end{center}
\vspace{-3mm}
\end{figure}

We present the classification rate curves of each domain and the average classification rate for various $\lambda_e$, where $\lambda_c=1$, in Figures \ref{fig:roc} and \ref{fig:avg}, respectively, to examine the results more closely. Figure \ref{fig:numofsource} shows the experimental result for the case where some parts of the data from the old domain can be accessed. This figure illustrates that our method was significantly more effective than the traditional fine-tuning method when the old-domain data were partially accessible.

\subsection{Realistic dataset (ImageNet)}
The second experiment was an experiment using an ImageNet 2012 dataset. This dataset was more realistic because the resolution of the training images was much higher than that in the other datasets, such as CIFAR-10, MNIST, and SVHN. The dataset also contained realistic scenarios, such as lighting changes and background clutter. We used a subset of the dataset and randomly chose 50 classes from the original 1000 classes to save training time. We also used image brightness to divide the images into old and new domains. The normal brightness images were put in the old domain, while relatively bright or dark images were put in the new domain. 

\begin{table}[h!]
\centering
\scriptsize
\caption{Experimental results for the realistic dataset.}
\label{googlenet}
\begin{center}
\begin{tabular}{c|cccc}
\hline\hline
& Methods&Old (\%)&New (\%)&Avg. (\%)\\ \hline
& Old network (ReLU) & 85.53 & 76.44 & 80.99 \\ \cline{2-5} 
Image & Fine-tuning (ReLU) & 80.06 & {85.74} & 82.90 \\ 
Net & Fine-tuning (Linear) & 80.16 & 84.36 & 82.26 \\ 
Normal& Modified LwF($\beta=2$) & 84.33 & 82.17 & 83.25 \\
$\cup$ & Modified LwF($\beta=1$) & 83.94 & 83.17 & 83.56 \\
Dark \&& Modified LwF($\beta=0.1$) & 80.46 & {85.54} & 83.00 \\\cline{2-5} 
Bright & \textbf{LF} ($\lambda_{e}=10^{-2}$) & {85.10} & 83.56 & 84.04 \\
& \textbf{LF} ($\lambda_{e}=5\times10^{-3}$) & 84.98 & 84.4 & \textbf{84.69} \\
& \textbf{LF} ($\lambda_{e}=10^{-3}$) & 83.92 & 84.75 & 84.34 \\
& \textbf{LF} ($\lambda_{e}=5\times10^{-4}$) & 83.05 & {85.54} & {84.30} \\ \hline\hline
\end{tabular}
\end{center}
\vspace{-3mm}
\end{table}

Table \ref{googlenet} shows the experimental results for the ImageNet dataset. The experimental results in the previous section clearly showed that the traditional fine-tuning technique has forgotten much about the old domain. Furthermore, it showed that the modified LwF can also mitigate the forgetting problem, and our method remembered more information from the old domain than the modified LwF. On average, our method improved the recognition rate by about 1.8\% compared to the existing fine-tuning method.
\section{Discussions}
\subsection{Are Maxout and LWTA activation functions helpful for mitigating the catastrophic forgetting?}
From the experimental result shown in Table \ref{tb:Result}, we conclude that the effect is not significant. Maxout showed the best performance, and LWTA showed a performance similar to that of ReLU. This might be caused by an increase of learnable parameters because Maxout uses additional parameters for learning piecewise activation functions. As a result, Maxout shows relatively low accuracy compared to state-of-the-art techniques, such as EWC, modified LwF, and our proposed method. This implies that simply changing activation functions is not very helpful in mitigating the catastrophic forgetting problem.

\textcolor{blue}{
}
\subsection{Limitation of the EWC}
Our experimental results showed the limitations of the EWC method. The problem emerges when some diagonal elements of the Fisher information matrix are very close to zero. In this case, even if the value of $\gamma$ is maximized, a forgetting problem will occur as $l_2$ loss does not work because of the extremely small values of the Fisher information matrix.

There is another problem arising from the fact that the Fisher information matrix is computed using the training data of the old domain. The Fisher information matrix is a key parameter to alleviate the catastrophic forgetting problem in the EWC method, and the matrix may be inaccurate to the test data of the old domain. Therefore it may fail on the test data of the old domain, and this makes the new network forgets a lot. 

\begin{figure}[t!]
\begin{center}
\includegraphics[width=4.1cm]{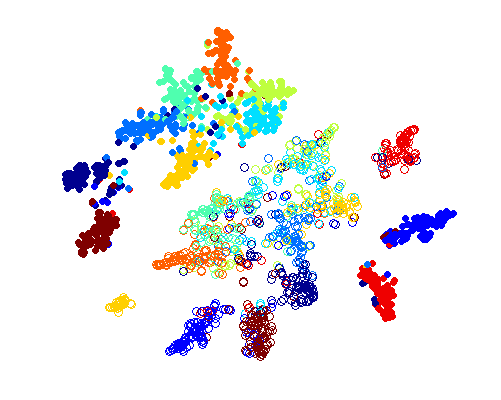}~
\includegraphics[width=4.1cm]{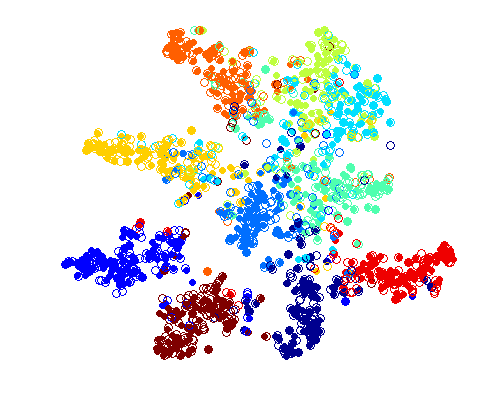}\\
\small
(a)\hspace{4.3cm}(b)
\vspace{-3mm}
\caption{\textbf{Visualization of the feature space for ten classes using t-SNE \cite{van2008visualizing}. \textit{[Best viewed in color]}} Each color represents each class. Filled circles denote features of the old training data extracted by the old network. Circles represent features of the old training data extracted by the new network. (a) Traditional fine-tuning method. (b) Proposed method.}
\vspace{-3mm}
\label{fig:feature}
\end{center}
\end{figure}

\begin{figure*}[t!]
\begin{center}
\includegraphics[width=5.8cm]{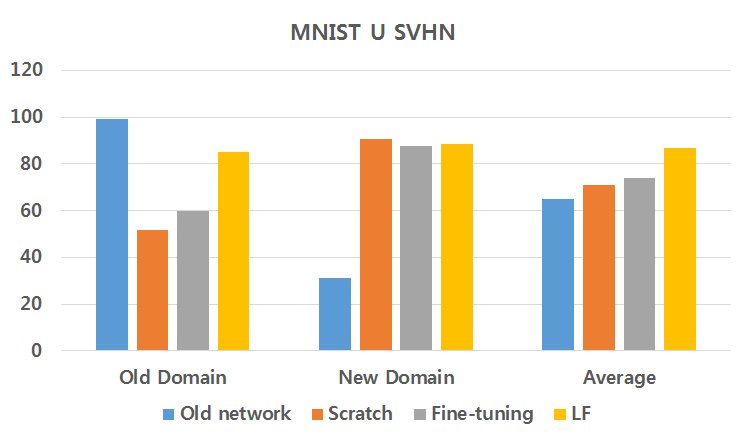}
\includegraphics[width=5.8cm]{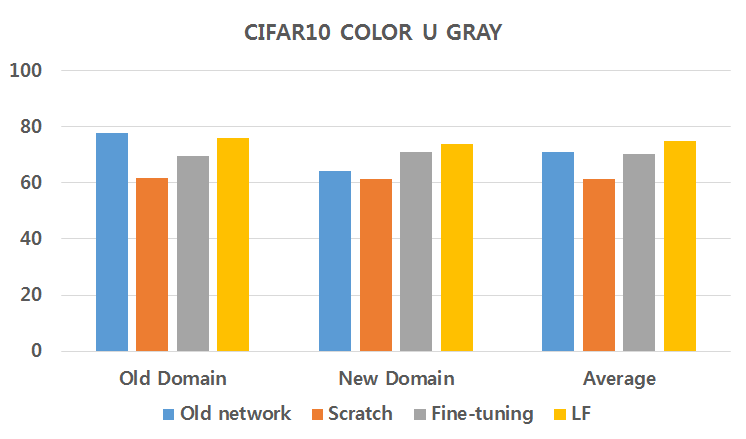}
\includegraphics[width=5.8cm]{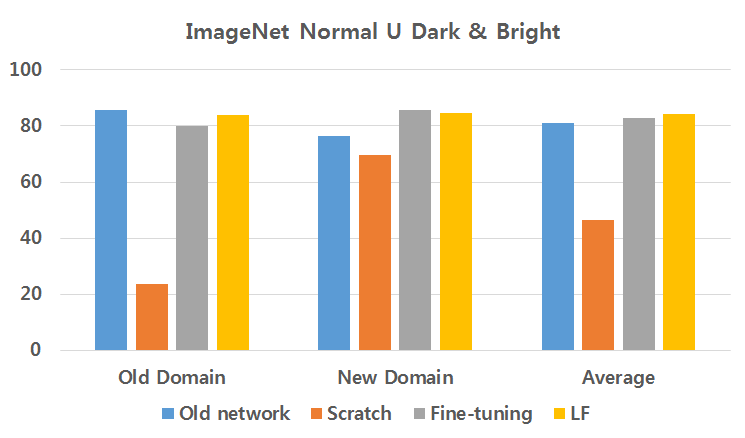}\\
(a)\hspace{5.5cm}(b)\hspace{5.5cm}(c)
\small
\vspace{-3mm}
\caption{\textbf{Scratch learning VS Fine-tuning VS LF learning on new domain. \textit{[Best viewed in color]}} Y-axis represents classification rates for each approach. ``Old network'' has been trained using the old domain training data. ``Scratch'' means a network trained only using the new domain training data from random weights. ``LF'' is our proposed method.}
\vspace{-3mm}
\label{fig:performance}
\end{center}
\end{figure*}

\subsection{Effectiveness of the LF}
Figures \ref{fig:feature} (a) and (b) show the feature spaces after the traditional fine-tuning method and our proposed method are executed, respectively. In the proposed method, high level features of old domain data, which are extracted by each network (old and new), are well clustered, even if re-training only using the new data is finished. Moreover, old domain features extracted from each network are well mixed, and they are not distinguishable from each other in the proposed method. This is probably due to the $\mathcal{L}_e$ loss, and it might prevent significant changes in the feature space.

\subsection{Further Analysis of Scratch learning, Fine-tunig and LF learning}
Additional experiment such as learning from scratch on new domain was conducted for further analysis.
First, we initialize neural networks from random weights and train them using only the data from the new domain, and we report a comparison of three different methods.

In the case of MNIST $\cup$ SVHN shown in Figure \ref{fig:performance} (a), a new network trained from scratch achieves the best performance in the new domain (indicated by orange color). On the other hand, the performance of the old domain is not good. This phenomenon is natural because the network did not see any old domain data. Furthermore, we observed that there is no improvement of the fine-tuning method for the new domain because the amount of data in both MNIST and SVHN is large enough to learn the new domain. The positive effect of fine tuning may occur when the number of new domain data is small as in the CIFAR Color $\cup$ Gray and ImageNet normal $\cup$ Dark \& Bright experiments. One interesting point in this experiment is that the average performance of ``Scratch'' (trained only using SVHN) for both domains is better than that of the ``Old network'' (trained using MNIST). From this observation, we infer that a network trained with more complex data will have better generalization performance on other domains.

In the CIFAR Color $\cup$ Gray experiment, the number of training images for each domain is different. The number of training images in the new domain is 10,000, and the number of training images in the old domain is 50,000. Training images of the new domain are a disjoint set of training images of the old domain converted into grayscale images. Interestingly, the network trained only using training images of the new domain does not show a performance gap between old and new domains, as shown in Figure \ref{fig:performance} (b). This means that weights computed from grayscale images are also useful for distinguishing color images. We also observe that the performance of the scratch learning on the new domain is significantly lower than that of the conventional fine-tuning method because the number of training images in the new domain is small.

In the case of ImageNet Normal $\cup$ Dark \& Bright experiment, the number of training images in the new domain is much smaller than that in the old domain (e.g. 52,503 vs 5,978). Similar to the CIFAR experiment, the fine-tuning method outperforms learning from the scratch on the new domain, as shown in Figure \ref{fig:performance} (c). Moreover, unlike the CIFAR experiment, the classification rate on the old domain of the scratch learning is the lowest among three different methods. In this case, we think that an overfitting problem occurred because there are few training images in the new domain.

\subsection{Feasibility for Continual Learning}
To show the feasibility of our algorithm for a continual learning problem, we conducted further experiments using the CIFAR-10 dataset.
Our experimental protocol is as follows. The CIFAR-10 dataset is manually separated into ten disjoint sets, and each group is input sequentially to the network. We assumed that previous groups are not accessible. Each group is trained during iteration $10,000$, and a total of $10,000~(\text{iterations}) \times 10~(\text{groups})= 100,000$ was used for both fine-tuning and LF learning. For the offline learning, we used $60,000$ iterations, and this method used whole training data sets.
From the results of Table \ref{result_inc}, we conclude that fine-tuning is not effective in the continual learning case, but our proposed LF method shows good results. As verified in the previous section, our method remembers the information of old data sets, and hence can achieve better results. From the result, we think that our LF method might be applied to the continual learning problem.

\begin{table}[h]
\footnotesize
\centering
\caption{Continual learning test.}
\label{result_inc}
\begin{center}
\begin{tabular}{c|c|c|c}
\hline\hline
& {Offline} & {Fine-tuning} & \bf{LF}\\ \hline
Classification rate & 78.16 & 67.18 & \textbf{71.1} \\ \hline\hline
\end{tabular}
\end{center}
\vspace{-3mm}
\end{table}

\section{Conclusion}
In this paper, we introduced a domain expansion problem and proposed a new method, called the less-forgetful learning, to solve the problem.
Our method was effective in preserving the information of the old domain while adapting to the new domain. Our method also outperformed other existing techniques such as fine-tuning with different activation functions, the modified LwF method, and the EWC method. In the experiments, our learning method was applied to the image classification tasks, but it is flexible enough to be applied to other tasks, such as speech and text recognition.



{
\bibliographystyle{aaai}
\bibliography{dnn}
}
\end{document}